%% file: paper-arxiv-corrected.tex
\documentclass[runningheads]{llncs}
\usepackage[T1]{fontenc}
\usepackage{graphicx}
\usepackage{booktabs}
\usepackage[misc]{ifsym}
\newcommand{\corr}{(\Letter)}
\usepackage{float}
\usepackage{subcaption,xcolor,lipsum}
\usepackage{tabularx}
\usepackage{multirow}
\newcolumntype{L}[1]{>{\hsize=#1\hsize\raggedright\arraybackslash}X}
\newcolumntype{C}[1]{>{\hsize=#1\hsize\centering\arraybackslash}X}
\newcolumntype{R}[1]{>{\hsize=#1\hsize\raggedleft\arraybackslash}X}
\usepackage{pifont}
\newcommand{\cmark}{\ding{51}}%
\newcommand{\xmark}{\ding{55}}%

\usepackage{amssymb, amsmath}
\usepackage{enumerate}
\usepackage{hyperref}
\usepackage[depth=2]{bookmark}

\begin{document}

\title{\textsc{ReDeLEx}: A Framework for \\Relational Deep Learning Exploration}

\titlerunning{Relational Deep Learning Exploration}

\author{
Jakub Pele\v{s}ka  \corr \and
Gustav \v{S}\'{\i}r 
}

\authorrunning{J. Pele\v{s}ka and G. \v{S}\'{\i}r}

\institute{
Czech Technical University in Prague,\\
Karlovo náměstí 13, Prague, 121 35, Czechia
\email{jakub.peleska@fel.cvut.cz,gustav.sir@cvut.cz}
}

\tocauthor{Jakub~Pele\v{s}ka, Gustav~\v{S}\'{\i}r}
\toctitle{\textsc{ReDeLEx}: A Framework for Relational Deep Learning Exploration}

\maketitle              

\begin{abstract}
Relational databases (RDBs) are widely regarded as the gold standard for storing structured information. Consequently, predictive tasks leveraging this data format hold significant application promise. Recently, Relational Deep Learning (RDL) has emerged as a novel para\-digm wherein RDBs are conceptualized as graph structures, enabling the application of various graph neural architectures to effectively address these tasks. However, given its novelty, there is a lack of analysis into the relationships between the performance of various RDL models and the characteristics of the underlying RDBs. 

In this study, we present \textsc{ReDeLEx}---a comprehensive exploration framework for evaluating RDL models of varying complexity on the most diverse collection of over 70 RDBs, which we make available to the community. Benchmarked alongside key representatives of classic methods, we confirm the generally superior performance of RDL while providing insights into the main factors shaping performance, including model complexity, database sizes and their structural properties.


\end{abstract}

\section{Introduction}

From their establishment~\cite{Codd1970}, {Relational Databases} (RDBs) played a pivotal role in transforming our society into the {current} information age. Data stored as interconnected tables, safeguarded by integrity constraints, have proven to be an effective method for managing domain information. 
Consequently, RDBs still prevail today as a backbone of critical systems in a number of important domains ranging from healthcare~\cite{white_pubmed_2020} to government~\cite{maali_enabling_2010}.

Although ubiquitous in modern application stacks, the data format of RDBs is deeply incompatible with classic Machine Learning (ML) workflows, which assume data in the standard form of fixed-size i.i.d. feature vectors, forming the common ``tabular'' learning format. Nevertheless, this assumption is clearly violated with the relationships between the differently-sized RDB tables. To address the discrepancy, the historically prevailing approach has been to turn the relational into the tabular format by means of ``\textit{propositionalization}''~\cite{propos}, which is essentially a feature extraction routine where relational substructures get aggregated from the relations into the attributes (features) of the tabular format, upon which classical ML methods may then operate. Nevertheless, this comes at the cost of information loss during this preprocessing step.

Recently, building on advances in graph representation learning~\cite{hamilton_graph_2020}, deep learning models directly exploiting the relational structure of RDBs have started to gain traction~\cite{Cvitkovic2020,zhang2023gfs,zahradnik2023deep,peleska_transformers_2024}, establishing the field of \textit{Relational Deep Learning} (RDL)~\cite{fey2024position}. Following the ``message-passing'' principles of \textit{Graph Neural Networks} (GNN;~\cite{wu2020comprehensive}), RDL models treat the structure of an RDB as a heterogeneous (temporal) graph, where individual table rows correspond to nodes, and edges are formed through integrity constraints set by the primary and foreign keys. Utilizing the graph representation then allows for the application of various GNNs, and their various extensions, with adapted message-passing schemes.

The generality and spread of RDBs allow for a broad spectrum of domain information to be stored, upon which a variety of predictive tasks can be formulated, each with unique aspects and qualities. This presents a challenge for establishing a broad enough benchmark to appropriately assess the general performance of RDL.
Currently, the most prominent effort in this area is the recently proposed \textsc{RelBench}~\cite{robinson2024relbench}, which introduced the evaluation of RDL, albeit with a very limited scope of simple models and just five accessible datasets. However, the overarching domain of \textit{relational learning}~\cite{Raedt}, currently ignored by the RDL community, has a rich history of working with the relational data format~\cite{muggleton1994inductive,cropper2020turning30}, including benchmarking of the propositionalization techniques~\cite{propos}. Notably, this includes the CTU Relational Learning Repository~\cite{motl2015ctu} that historically collected more than 70 diverse RDBs.

Our aim in this paper is to provide a bridge between the communities of traditional (logic-based) relational learning~\cite{Raedt} and the contemporary RDL~\cite{robinson2024relbench} towards a more comprehensive evaluation of the diverse existing methods.
To that aim, we introduce \textsc{ReDeLEx}---an experimental framework for developing and benchmarking diverse RDL architectures against classic methods over the most comprehensive collection of tasks and datasets to date. The implementation of the framework is readily available on GitHub.\footnote{\url{https://github.com/jakubpeleska/ReDeLEx}}

\section{Background}

In this paper, we experimentally explore learning from RDBs (Sec.~\ref{sec:rdb}) with GNN-based models (Sec.~\ref{sec:gnn}) resulting in the RDL methodology (Sec.~\ref{sec:rdl}).

\subsection{Relational Databases}
\label{sec:rdb}

Principles of RDBs are formally based on the \textit{relational model}~\cite{codd1990relational}, which is grounded in relational logic~\cite{gallier2015logic}. This abstraction enables the definition of any database, regardless of specific software implementation, as a collection of $n$-ary relations, which are defined over the domains of their respective attributes, managed by the Relational Database Management System (RDBMS) to ensure data consistency with the integrity constraints of the database schema.
The key concepts to be used in this paper are as follows.

\subsubsection{Relational Database} A Relational Database (RDB) $\mathcal{R}$ is defined as a finite set of relations $R_1, R_2, \dots ,R_n$. An instance of an RDB $\mathcal{R}$ is implemented through a RDBMS, enabling to perform Structured Query Language (SQL;~\cite{chamberlin_sequel_1974}) operations, rooted in {relational algebra}.

\subsubsection{Relation (Table)} Formally, an $n$-ary relation $R_{/n}$ is a subset of the Cartesian product defined over the domains $D_i$ of its $n$ \textit{attributes} $A_i$ as $R_{/n} \subseteq D_1 \times D_2 \times \dots \times D_n$, where $D_i = \mathsf{dom}(A_i)$. Each relation $R$ consists of a heading (signature) $R_{/n}$, formed by the set of its attributes, and a body, formed by the values of the respective attributes, commonly represented as a \textit{table} $T_R$ of the relation $R$.

\subsubsection{Attribute (Column)} \textit{Attributes} $\mathcal{A}_R = \{A_1, \ldots, A_n\}$ define the terms of a relation $R_{/n}$, corresponding to the \textit{columns} of the respective table $T_R$. Each attribute is a pair of the attribute's name and a \textit{type}, constraining the domain of each attribute as $\mathsf{dom}(A_i) \subseteq \mathsf{type}(D_i)$. An attribute \textit{value} $a_i$ is then a specific valid value from the respective domain of the attribute $A_i$.

\subsubsection{Tuple (Row)} An $n-$\textit{tuple} in a relation $R_{/n}$ is a tuple
of attribute values ${t_i} = (a_1, a_2, \ldots, a_n)$, where $a_j$ represents the value of the attribute $A_j$ in $R$. The relation can thus be defined extensionally by the \textit{unordered} set of its tuples: $R = \{t_1, t_2,\ldots, t_m\}$, corresponding to the \textit{rows} of the table $T_R$.

\subsubsection{Integrity constraints} In addition to the domain constraints $\mathsf{dom}(A_i)$, the most important integrity constraints are the primary and foreign keys. A \textit{primary} key $PK$ of a relation $R$ is a minimal subset of its attributes $R[PK] \subseteq \mathcal{A_R}$ that uniquely identifies each tuple: 
$\forall t_1, t_2 \in R:~ (t_1[PK] = t_2[PK]) \Rightarrow (t_1 = t_2)$. 
A~\textit{foreign} key ${FK}_{R_2}$ in relation $R_1$ then refers to the primary key ${PK}$ of another relation $R_2$ as 
$\forall t \in R_1:~ t[FK] \in \{t'[PK] \mid t' \in R_2\} \,.$
This constitutes the inter-relations in the database, with the RDBMS handling the \textit{referential integrity} of ${T_{R_1}}[FK] \subseteq {T_{R_2}}[PK]$.

\subsection{Graph Neural Networks}
\label{sec:gnn}
Graph Neural Networks constitute a comprehensive class of neural models designed to process graph-structured data through the concept of (differentiable) \textit{message-passing}~\cite{wu2020comprehensive}. Given an input graph $G = (\mathcal{V}, \mathcal{E})$, with a set of nodes $\mathcal{V}$ and edges $\mathcal{E}$, let $h_v^{(l)} \in \mathbb{R}^{d^{(l)}}$ be the vector representation (embedding) of node $v$ at layer $l$.
The general concept of GNNs can then be defined through the following sequence of three functions:

\begin{enumerate}
    \item[(i)] \textit{Message} function $M^{(l)}: \mathbb{R}^{d^{(l)}} \times \mathbb{R}^{d^{(l)}} \to \mathbb{R}^{d_m^{(l)}}$ computes messages for each edge $(u, v) \in E$ as 
        $m_{u \to v}^{(l)} = M^{(l)}(h_u^{(l)}, h_v^{(l)}) \,.$
    
    \item[(ii)] \textit{Aggregation} function $A^{(l)}: \{\mathbb{R}^{d_m^{(l)}}\} \to \mathbb{R}^{d_m^{(l)}}$ aggregates the messages for each $v \in V$ as 
       $M_v^{(l)} = A^{(l)}\left(\{m_{u \to v}^{(l)} ~|~ (u, v) \in E\}\right) \,.$
    
    \item[(iii)] \textit{Update} function $U^{(l)}: \mathbb{R}^{d^{(l)}} \times \mathbb{R}^{d_m^{(l)}} \to \mathbb{R}^{d^{(l+1)}}$ updates representation of each $v \in V$ as 
        $h_v^{(l+1)} = U^{(l)}(h_v^{(l)}, M_v^{(l)}) \,.$
\end{enumerate}

The specific choice of message, aggregation, and update functions varies across specific GNN models, which are typically structured with a predefined number $L$ of such layers, enabling the message-passing to propagate information across $L$-neighborhoods within the graph(s).

\subsection{Relational Deep Learning}
\label{sec:rdl}
In this paper, we adopt the concept of RDL as extending mainstream deep learning models, particularly the GNNs (Sec.~\ref{sec:gnn}), for application to RDBs (Sec.~\ref{sec:rdb}). For completeness, in the relational learning community~\cite{cropper2020turning30}, a number of similar approaches combining relational (logic-based) and deep learning methods arose under a similar name of ``deep relational learning''~\cite{vsir2021deep}. Nevertheless, for compatibility with the recently introduced frameworks~\cite{fey2024position}, we hereby continue with the contemporary RDL view, where RDBs are first transformed into a graph-based representation suitable for the GNN-based learning. 

\subsubsection{Database Representation} 
\label{sec:db-graph}

The fundamental characteristic of RDL~\cite{fey2024position} is to represent an RDB as a heterogeneous graph.\footnote{sometimes referred to as the ``relational entity graph''} 
The graph representation can be defined as $G = (\mathcal{V}, \mathcal{E}, \mathcal{T}^v, \mathcal{T}^e)$, where $\mathcal{V}$ is the set of nodes, $\mathcal{E}$ is the set of edges, $\mathcal{T}^v$ is a set of node types with a mapping $\phi: \mathcal{V} \to \mathcal{T}^v$, and  $\mathcal{T}^e$ is a set of edge types with a mapping $\psi: \mathcal{E} \to \mathcal{T}^e$. The node types and edge types collectively form the graph \textit{schema} $(\mathcal{T}^v, \mathcal{T}^e)$.

Given an RDB schema $\mathcal{R}$, the node types $T \in \mathcal{T}^v$ correspond to the relations (tables) $T$ within the database $\mathcal{T}^v \overset{1:1}{\to} \mathcal{R}$,
while the edge types $\mathcal{T}^e$ represent the undirected inter-relations between the tables, as defined by the primary-foreign key pairs:
$\mathcal{T}^e = \{({R_i, R_j})~|~{R_i}[FK_{R_j}] \subseteq {R_j}[PK]~\lor~{R_j}[FK_{R_i}] \subseteq {R_i}[PK]\} \text{.}$
For a specific \textit{instance} of an RDB $\mathcal{R}$, the set of nodes $\mathcal{V}$ is then defined as the union of all tuples (rows) $t_i$ from each relation
$\mathcal{V} = \{v_{i,j}~|~R_i \in \mathcal{R},~t_j \in R_i\} \text{,}$
and the set of edges $\mathcal{E}$ is defined as $\mathcal{E} = \{({v_{i,k}, v_{j,l}}) |~t_k \in R_i,~t_l \in R_j, (R_i, R_j) \in \mathcal{T}^e\}$.

The graph representation is further enriched by \textit{node embedding matrices}, \textit{attribute schema}, and optionally a \textit{time mapping}. {Node embedding matrix} $h^{(l)}_v \in \mathbb{R}^{d \times d_{\phi(v)}}$ contains the embedding representation of a node $v \in \mathcal{V}$ in a given layer $l$. With an {attribute schema} $\mathcal{A}_T$ that provides information about the types of attributes $A_1,\dots,A_n$ associated with the nodes $v$ of a specific node type $T \in \mathcal{T}^v$, the initial embedding tensors $h_v^{(0)} \in \mathbb{R}^{d^{(0)} \times n}$ are computed from the raw database attribute tuples ${t_i} = (a_1, a_2, \ldots, a_n)$ through multi-modal attribute encoders~\cite{fey2024position}. Finally, the \textit{time mapping} is a function $\tau$ that assigns a timestamp $t_v$ to each node $\tau: v \mapsto t_v $, effectively creating a dynamically growing graph in time, enabling the use of temporal graph sampling~\cite{rossi2020temporal}.

\subsubsection{Predictive Tasks}
In RDL, predictive tasks are implemented through the creation of dedicated training tables $T_t$ that extend the existing relational schema of $\mathcal{R}$. As introduced in~\cite{fey2024position}, a training table $T_t$ contains two essential components: foreign keys $T_t[FK]$ that identify the entities of interest and target labels $y \in \mathcal{A}_{T_t} \setminus T_t[FK]$. Additionally, timestamps $t_v \in \mathcal{A}_{T_t}$ that define temporal boundaries for the prediction of $y$ can also be included.

The training table methodology supports a diverse range of predictive tasks, including node-level predictions (e.g., customer churn, product sales), link predictions between entities (e.g., user-product interactions), and, crucially, both temporal and static predictions. In the case of temporal predictions, a timestamp attribute $t_v$ in the training table $T_t$ specifies when the prediction is to be made, restricting the model to only consider information available up to the point $t_v$ in time.

\subsubsection{Neural Architecture Space} 
\label{sec:rdl-neural-space}
Building upon the heterogeneous graph representation $G$, RDL models generally consist of the following four major stages.

\begin{enumerate}
\item \textbf{Table-level attribute encoder} creates the initial node embedding matrices $h_v^{(0)} \in \mathbb{R}^{d^{(0)} \times n}$, i.e. sequences of $n$ embedding vectors $\mathbb{R}^{d^{(0)}_{\phi(v)}}$ for each attribute $A_1,\dots,A_n$ of $\phi(v)$ based on its respective semantic data type.

\item \textbf{Table-level tabular model} allows to employ existing tabular learning models~\cite{chen2023trompt,hu2020tabtransformer} to yield more sophisticated node embeddings $h_v^{(l)}$. Notably, in this stage, an RDL model \textit{may} reduce the dimensionality of the node attribute matrix embedding $h_v^{(l)} \in \mathbb{R}^{d^{(l)} \times n}$ to a vector embedding $h_v^{(l)} \in \mathbb{R}^{d^{(l)}_{\phi(v)}}$.

\item \textbf{Graph neural model} then depends on the chosen embedding dimensionality of $h_v^{(l)}$. If there is a single embedding vector $h_v^{(l)} \in \mathbb{R}^{d^{(l)}_{\phi(v)}}$ per each node, the model can employ standard GNN~(Sec.~\ref{sec:gnn}) heterogeneous message-passing~\cite{velivckovic2018graph,brody2022how}, otherwise a custom message-passing scheme~\cite{peleska_transformers_2024} is required.

\item \textbf{Task-specific model head} finally provides transformation of the resulting node embeddings into  prediction, usually involving simple MLP layers.
\end{enumerate}

\section{The \textsc{ReDeLEx} Framework}
The Relational Deep Learning Exploration (ReDeLEx) framework, which we introduce in this paper, offers a comprehensive environment for evaluating various RDL architectures over diverse RDB datasets.

\subsection{Workflow Components}
\begin{figure}[t]
    \centering
    \includegraphics[width=\textwidth]{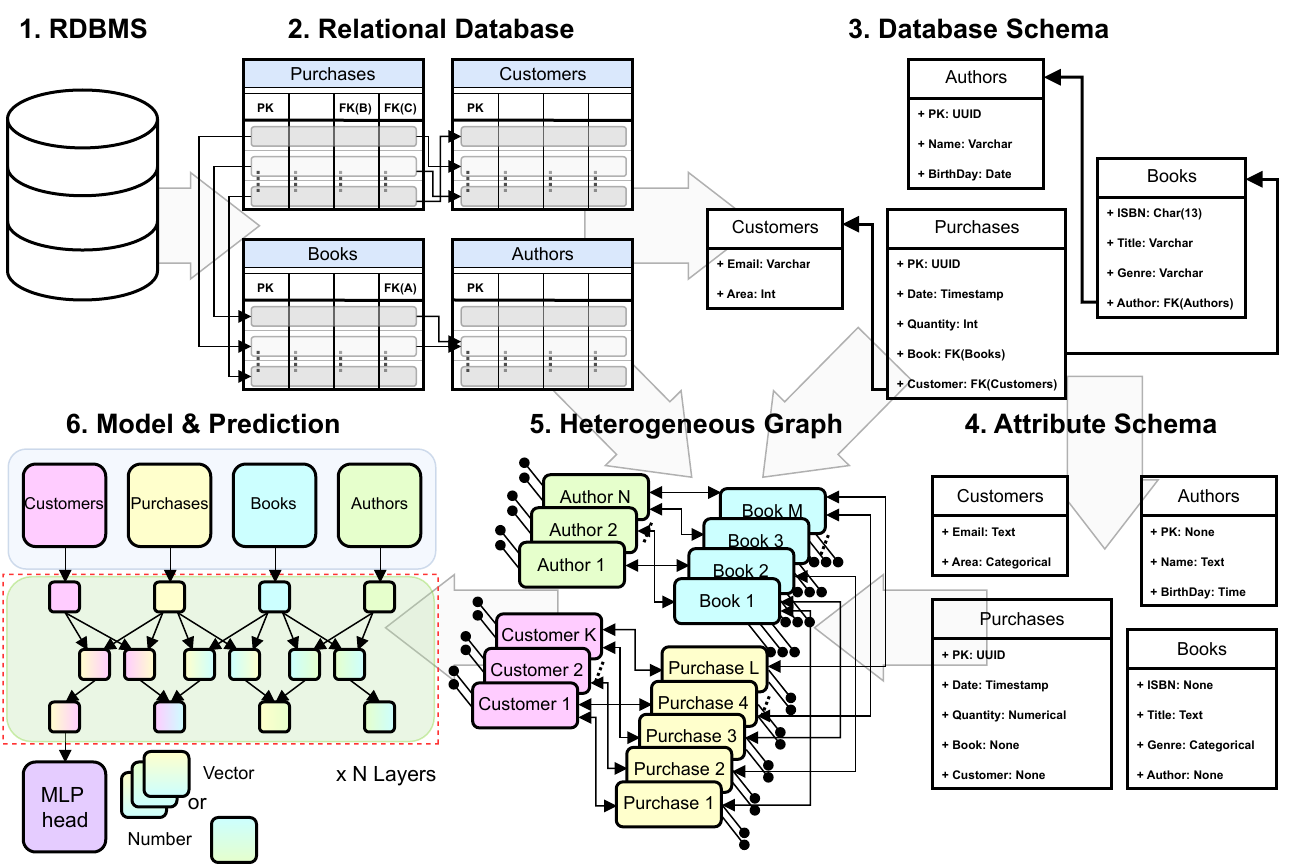}
    \caption{\textsc{ReDeLEx} end-to-end workflow for RDL.}
    \label{fig:pipeline}
\end{figure}

The ReDeLEx workflow, depicted in Fig.~\ref{fig:pipeline}, consists of modular blocks that enable systematic exploration of the neural architecture and database configuration space, significantly extending the current scope~\cite{robinson2024relbench} of RDL experimentation.


\subsubsection{Database Connectivity} 
In contrast to~\cite{robinson2024relbench}, the framework provides a standardized interface for connecting \textit{directly} to an RDB~\cite{sqlalchemy}, supporting various dialects of RDBMS. Notably, this enables a truly end-to-end deep learning pipeline, connecting to a possibly \textit{remote} RDBMS, hosting the target RDB.

\subsubsection{Attribute Schema}
Attribute schema creation, which is a crucial yet overlooked step in RDL, mediates information regarding the attribute types $A_T$ within the specific node type $T \in \mathcal{T}^v$ based on the original table attributes $A \in \mathcal{A}_T$. \textsc{ReDeLEx} automatically generates the attribute schema based on the SQL types and data from the RDBMS. Note that assessing a semantic type $\mathsf{dom}(A_i)$ is not straightforward since, e.g., a SQL \texttt{VARCHAR} attribute $A_i$ often stores categorical, textual, as well as temporal values $a_i$. To disambiguate such cases, we employ in-built heuristics utilizing the SQL types, names of the attributes, ratio of unique values, and patterns in the data to facilitate proper attribute embedding.

\subsubsection{Predictive Tasks}
The existing benchmark~\cite{robinson2024relbench} provides support solely for tasks with a training table $T_t$ generated from historical data through an SQL query. While useful, without any changes to the underlying database, this setting renders many RDB prediction tasks infeasible. \textsc{ReDeLEx} addresses this problem by adding support for tasks that require more substantial modifications of the original database. Tasks leveraging this functionality then not only generate a new table $T_t$ but a whole modified instance $\mathcal{R}'$ of the original database $\mathcal{R}$. 

For example, assume the most common case where the database $\mathcal{R}$ already contains the target attribute $A_T$, used for some node-level prediction task. In such a case the table $T_t$ containing the target needs to be split into two tables $T_{t_1}, T_{t_2}$ where $T_{t_1}$ contains all original data except the target attribute $A_T$ and is part of the newly modified database $\mathcal{R}'$, and $T_{t2}$ contains a duplicate of the primary key $T_t[PK]$, now used as a foreign key to the original table $T_t$, and the target attribute $A_T$. This table $T_{t2}$ is then used as the new training table $T_t'$.

Importantly, this scheme can be applied to generate tasks for \textit{unsupervised pretraining}. Pretraining tasks can be created by choosing any table $T \in \mathcal{R}$ in the database and duplicating it as $T'$. The unchanged duplicate $T'$ can then be used as a training table $T_t$, while the values of cells in the original table $T$ are randomly removed (masked out). The task is then to reconstruct any missing values in the classical tabular learning fashion~\cite{arik2021tabnet}, opening possibilities for sophisticated pretraining methods~\cite{somepalli2021saint}.

\subsection{RDL-Suitable Databases and Tasks}
Due to the generality of the relational model, RDBs often contain data with vastly diverse structural characteristics that, in some cases, do not properly exploit the relational model (Sec.~\ref{sec:rdb}). Likewise, not all of the 70+ available RDBs~\cite{motl2015ctu} are actually suitable for the relational learning models. In this section, we examine RDB characteristics in the context of RDL to identify suitable databases to be used in the experiments~(Sec.~\ref{sec:experiments}).

\subsubsection{Database Characteristics}
\label{sec:db-character}
To assess their overall characteristics, \textsc{ReDeLEx} associates each database task with various features pertinent to different parts if the training workflow (Fig.~\ref{fig:pipeline}), which can be split into the below categories.

\begin{enumerate}
    \item \textbf{Database features} provide high-level view of the data, including a domain (e.g. medicine, government, sport), whether the database is artificial or not, number of tables inside the database, number of foreign keys, number of factual (non-key) columns, number of columns with a specific variable type (e.g. numerical, categorical, time), total number of rows and total number of primary-foreign key pairs.
    
    \item \textbf{Schema features} describe high-level structural aspects of the data. This includes the multiplicity of the relationships between the tables (one-to-one, one-to-many, many-to-many), features of the undirected graph induced by the primary-foreign key pairs (e.g. graph diameter,\footnote{Graph diameter is the maximum length of all the shortest paths between the nodes.} or cycle detection).
    
    \item \textbf{Task features} provide a similar type of information as the database features that are specific to the task and its target entity tables. This includes whether the task is temporal or static, number of training samples, multiplicity of relationships of the target entity table, etc.
    
    \item \textbf{Graph features} inform about the properties of the transformed heterogeneous graph including, e.g., average eccentricity\footnote{The eccentricity of a node is the maximum distance from the node to all other nodes.} of nodes or graph density.
\end{enumerate}

\subsubsection{Tabular Data}
\label{sec:tab-like}
A salient feature of RDBs are the inter-relations between the tables (Sec.~\ref{sec:rdb}). As such, it is obvious that RDBs that contain a single table, or multiple tables without any primary-foreign key pairs, will not benefit from the use of RDL. Furthermore, databases consisting of multiple tables linked solely by one-to-one relationships fall under the same category, as they allow for a complete \textit{join} of the whole RDB into a single table.
Importantly, as all values of foreign keys are unique (with the exception of missing values), all the resulting rows remain independent of each other, turning the RDL setting into standard tabular learning (see App. Tab.~\ref{tab:tab-like-info} for a list of such databases).

\subsubsection{Graph Data}
\label{sec:graph-like}
On the other hand, RDBs are also characterized by building on the tabular representation, where an arbitrary number of attributes can be connected by a single relation. This is in contrast to the graph data which correspond to \textit{binary} relational structures. Consequently, natively graph-structured data, such as molecules or family trees, although possible to be stored in an RDB, also do not fully exploit the relational model. In such cases, the RDL paradigm reduces to the simpler GNN setting~\cite{hamilton_graph_2020}, introducing an unnecessary complexity otherwise. More generally, RDL models for tasks on RDBs with a low number of non-key attributes (see Sec.~\ref{sec:experiment3}) may suffer from information sparsity (see App. Table~\ref{tab:graph-like-info} for databases with the stated characteristics.)

\section{Experiments}
\label{sec:experiments}

The aim of the experiments presented in this section is to demonstrate \textsc{ReDeLEx} in exploring the following selected RDL research questions:

\begin{enumerate}[{Q}1:]
    \item How do RDL methods perform in comparison to the traditional methods over diverse benchmarking tasks (Sec.~\ref{sec:experiment1})?
    \item Is it possible to apply tabular learning to a non-trivial RDB task while achieving results comparable to the RDL methods (Sec.~\ref{sec:experiment2})?
    \item What are some of the essential RDB characteristics that contribute to a successful application of a given learning model (Sec.~\ref{sec:experiment3})?
\end{enumerate}

\subsubsection{Databases}
To establish a comprehensive yet manageable list from the overall 70+ available RDBs~\cite{motl2015ctu} for the RDL experimentation, we separated databases that exhibit the tabular (Sec.~\ref{sec:tab-like}) or graph-like (Sec.~\ref{sec:graph-like}) characteristics, or are artificially\footnote{with a single exception of the \texttt{tpcd} database} created (see App. Table~\ref{tab:db-info} for the most suitable databases).

\subsubsection{RDL Models}
\label{sec:rdl-models}
\textsc{ReDeLEx} is designed to accommodate development of highly diverse RDL architectures. For comprehensibility of the experiments, we present three models of gradually increasing complexity, selected from recent works. All the models fit into the outlined neural architecture space (Sec.~\ref{sec:rdl-neural-space}), while utilizing the same attribute encoders for the numerical, categorical, multi-categorical, textual, and temporal values.

\begin{enumerate}
    \item \textbf{GraphSAGE with Linear Transformation} is the simplest of the RDL models, applying a linear transformation on top of a concatenation of the attribute $a_1,\dots,a_n$ embeddings $h_v^{(0)} \in \mathbb{R}^{n \cdot d^{(0)}}$ to yield a single embedding vector $h_v^{(1)} = {W} h_v^{(0)}$ for each node $v$.
    The projected node embeddings $h^{(1)} \in \mathbb{R}^{d_{\phi(v)}}$ then form input into the GraphSAGE~\cite{hamilton2017inductive} model, forming the GNN stage. Finally, a task specific model head is applied.
    \item \textbf{GraphSAGE with Tabular ResNet} is similar to the previous, with the tabular-level stage reducing the node embedding dimensionality, however, the operation is performed through a more sophisticated tabular ResNet model~\cite{gorishniy_revisiting_2021}. 
    Notably, this model was previously used in~\cite{robinson2024relbench}, allowing to directly align results between the \textsc{RelBench} and \textsc{ReDeLEx} benchmarks.
    \item \textbf{\textsc{DBFormer}} is an implementation of the Transformer-based RDL model from~\cite{peleska_transformers_2024}. In constrast to the previous,\footnote{The key difference can be viewed analogously to the ``fusion'' and ``cooperation'' in multi-modal learning~\cite{hu2021unit,liang2024foundations}, considering the attributes as modalities. While the first two models \textit{fuse} the representations of the discrete attributes at the beginning, the \textsc{DBFormer} allows for \textit{cooperation} of the attributes through the GNN stage.} the model retains the original node embedding dimensionality $h_v^{(l)} \in \mathbb{R}^{n \times d^{(l)}}$ while exploiting the attention mechanism~\cite{vaswani2017attention} for learning interactions between both the attributes and tuples through a custom message-passing scheme.
\end{enumerate}

\subsubsection{Classical Models}
\label{sec:classic-models}
In addition to the selected RDL models, we include key representatives from related ML domains, including Gradient Boosted Decision Trees (GBDT;~\cite{natekin2013gradient}), Deep Tabular Learning (DTL;~\cite{borisov2022deep}) and Propositionalization (Prop.;~\cite{kramer2001propositionalization}).
Particularly, we compare against the LightGBM~\cite{ke_lightgbm_2017}---representative of GBDT; the getML's~\cite{getml} FastProp feature generator combined with XGBoost~\cite{chen2016xgboost}---representative of propositionalization; and the standalone tabular ResNet~\cite{gorishniy_revisiting_2021}---representative of deep tabular learning. Importantly, the LightGBM and the ResNet have access only to data from the task's target table, as these models fall into the tabular learning category. In contrast, the propositionalization method of FastProp with XGBoost exploits the full RDB structure.

\begin{table}[p]
    \centering
    \include{tables/class-overall-corrected}

    \caption{Overall results from the classification tasks, presenting AUC ROC values for the binary classification, and macro f1 score for the multiclass classification, respectively (higher is better). Static (non-temporal) tasks are tagged as ``orig.''}
    \label{tab:class-overall}
\end{table}

\subsection{Benchmarking tasks}
\label{sec:experiment1}
We present comprehensive results over two types of node-level tasks---binary classification and multiclass classification. The tasks can be further differentiated by the origin of the target labels and usage of temporal values. Tasks performed on datasets from the \textsc{RelBench} collection use generated target table attributes, while tasks on datasets from the CTU Relational use existing target table attributes. Additionally, tasks from the CTU Relational can be both static and temporal, while tasks from the \textsc{RelBench} collection are always temporal. 
Static and temporal tasks differentiate based on the constrains forced upon the sampling algorithm while generating a sub-subgraph used for training the model, and by the method of splitting the dataset between the training, validation and test data. Static tasks use neighborhood sampling~\cite{hamilton2017inductive} constrained only by the maximum number of neighbors, and data splitting is carried out at random w.r.t. a given ratio (e.g. 70:15:15). In contrast, temporal tasks extend the neighborhood sampling by incorporating temporal constraints, assuming only directed edges from nodes with an older timestamp and, similarly, the splits are carried out w.r.t. the timestamps where all training entities must precede validation and testing data, respectively (see App.~\ref{app:setup} for the full experimental setup).

The results on classification tasks shown in Table~\ref{tab:class-overall} demonstrate a strong performance of the RDL models (Sec.~\ref{sec:rdl-models}) over the classical models (Sec.~\ref{sec:classic-models}) on majority of the datasets. Specifically, the ResNet SAGE performs very well on binary classification tasks, especially on tasks with a training table generated from historical data. Nonetheless, both \textsc{DBFormer} and Linear SAGE perform just marginally worse compared to ResNet SAGE on average, showcasing the general robustness of the RDL paradigm. Notably, on the \texttt{hepatits}, \texttt{mondial}, \texttt{student loan}, \texttt{accidents}, and \texttt{genes} datasets, all the RDL models present near-perfect predictions while the classical models show an order of magnitude worse score, highlighting the contribution of the RDL representation.

\begin{table}[t]
    \centering
    \include{tables/class-tabular-corrected}
    \caption{Classification tasks over a subset of datasets formed by joining the target table, showing AUC ROC values for binary classification, and macro f1 score for multiclass classification, respectively (higher is better). Significant improvements (more than 0.05 score) are shown in bold, while new best results are underlined.}
    \label{tab:class-join}
\end{table}

\subsection{Tabular Learning}
\label{sec:experiment2}

In this scenario, we compare results of the Tabular Learning (TL) models from the previous section to ones trained on new tables formed by join operations over the RDBs (Sec.~\ref{sec:tab-like}). Particularly, we evaluate the TL models on tables generated by joins over \textit{both} one-to-one and many-to-one relationships of the 
target
table.\footnote{This is similar to a recent RelGNN method~\cite{chen_relgnn_2025}, albeit limited to the target table.}
Additionally, we include evaluation of RDL models with exactly $2$ layers in the graph neural stage, which is conceptually equivalent to the join operation. Finally, we include the overall best RDL models to put the results into context.
Note that the previous Table~\ref{tab:class-overall} demonstrated that the TL methods perform significantly worse on an absolute majority of tasks. In this experiment, we aim to assess whether a simple RDB transformation could actually change the situation in some cases.
Particularly, we select a subset of tasks from Table~\ref{tab:class-overall} where both the TL models showed at least $0.1$ worse score than the best RDL model. The results in Table~\ref{tab:class-join} show that, indeed, on a number of datasets the TL models register a significant improvement with results sometimes comparable to the best of RDL. Notably, on the \texttt{ergastf1}, \texttt{ncaa} and \texttt{tpcd} they even set new best results. This experiment demonstrates existing weak spots in the new RDL approach~\cite{fey2024position}, suggesting that caution and thorough analysis are still in order before deploying RDL on an RDB task.

\subsection{Essential Characteristics}
\label{sec:experiment3}
\begin{table}[t!]
    \centering
    \include{tables/database-characteristics-corrected}
    \caption{Characteristics of databases and their tasks selected based on the best performing model. Features are sorted into the categories described in Sec.~\ref{sec:db-character}.} 
    \label{tab:db-character}
\end{table}

The overarching aim of \textsc{ReDeLEx} is to assess common characteristics of RDBs and tasks in the context of a used learning approach. While a fully comprehensive assessment is out of scope of this short paper, in Table~\ref{tab:db-character} we summarize characteristics of the databases against the respective performances of the various model types. Following the analysis, RDL models generally tend to perform well on datasets with a large number of training samples and links. Propositionalization achieves best results mostly on smaller datasets, yet with a large number of factual (non-key) columns. This is in line with some previous studies~\cite{Lavrač2021}, despite the remaining prevalence of propositionalization methods in practice~\cite{getml}.
The TL methods then tend to perform best when there is a higher number of factual columns in the target table, which aligns with natural intuition. Moreover, these allow capturing more diverse attribute types where, e.g., both LightGBM and deep TL models are capable of utilizing textual attributes.

\subsubsection{Related Work}

As outlined in the Introduction, the \textsc{ReDeLEx} framework builds upon the CTU relational dataset collection~\cite{motl2015ctu} which it integrates with the \textsc{RelBench}~\cite{robinson2024relbench} interface to facilitate a wider scope of RDL~\cite{fey2024position} experimentation. As such, it is naturally related to recent works introducing new RDL models, which include~\cite{peleska_transformers_2024,zahradnik2023deep,chen_relgnn_2025}. Besides RDL, related work includes other dataset and benchmarking frameworks that address some facets of learning from relational data, including \cite{vogel2024wikidbs,wang_4dbinfer_2024}. The most salient feature of \textsc{ReDeLEx}, within the context of related work, is the provided bridge between the traditional relational learning methods~\cite{Raedt} and the contemporary RDL~\cite{fey2024position}.

\section{Conclusion}
In this study, we introduced \textsc{ReDeLEx}---a framework for exploring and evaluating Relational Deep Learning (RDL) models across diverse relational database contexts. The framework enables benchmarking on more than 70 databases, facilitating new insights into the relationships between the RDL neural architecture choices, traditional learning methods, and the underlying database characteristics.
Our results demonstrated that RDL approaches mostly outperform the traditional methods. Nevertheless, a closer inspection revealed important cases in which the performance of the competing tabular learning methods could be easily improved to match or even surpass RDL, highlighting the interim immaturity of the field, and the need for further RDL exploration. Our general exploration in this paper demonstrated that RDL performs well on databases with complex relationships and large numbers of samples, while the traditional methods may still remain a sensible choice for smaller and flatter datasets.

\begin{credits}
\subsubsection*{Ethical Considerations}
The performance of RDL models demonstrated in our research could enable more sophisticated inference of personal information from interconnected data sources. The framework's flexibility could facilitate deployment in domains with significant ethical implications, such as financial services, healthcare, and government operations. We encourage researchers using \textsc{ReDeLEx} to carefully assess privacy implications and implement appropriate anonymization techniques.

\subsubsection{\ackname} 
This work has received funding from the European Union’s Horizon Europe Research and Innovation program under the grant agreement TUPLES No. 101070149; and Czech Science Foundation grant No. 24-11664S.

\subsubsection{\discintname}
The authors have no competing interests to declare that are relevant to the content of this article.

\end{credits}
%
%
%
\bibliographystyle{splncs04}
\bibliography{bibliography}

\appendix
\section{Experimental Setup}
\label{app:setup}
All deep learning models (including RDL) were trained for a minimum of 10 epochs with a total minimum of 1000 steps of Adam~\cite{Kingma2014} optimizer with a learning rate of 0.001. Experiments on the RDL models were conducted with fixed hyperparameters with two exceptions---the number of layers of the graph neural model and the neighborhood graph sampling rate, which were searched for in a grid hyperparameter optimization. The hyperparameters include batch size---set to 512, message-passing aggregation function---set to summation, embedding vectors dimension, which is the same for both row and attribute embedding vectors---set to 64, neighborhood sampling rate---iterated over the values of 16, 32 and 64, and number of message-passing layers---a value in range of 1 to 4.

\section{Additional tables}
Here we provide additional tables with descriptive information about the data\-bases available through \textsc{ReDeLEx}.

\begin{table}[h]
    \centering
    \include{tables/tab-like}
    \caption{Tabular-like databases available through \textsc{ReDeLEx}.}
    \label{tab:tab-like-info}
\end{table}
\begin{table}[h]
    \centering
    \include{tables/graph-like}
    \caption{Graph-like databases available through \textsc{ReDeLEx}.}
    \label{tab:graph-like-info}
\end{table}

\begin{table}[p]
    \centering
    \include{tables/db-info}
    \caption{List of databases available for benchmarking in \textsc{ReDeLEx}.}
    \label{tab:db-info}
\end{table}

\end{document}

%% file: tables/class-overall-corrected.tex
\begin{tabular}{lc||cc|cc|cc|cc|cc|cc}
& &  \multicolumn{2}{c|}{GBDT} &  \multicolumn{2}{c|}{Prop.}  & \multicolumn{2}{c||}{Tab. DL} & \multicolumn{6}{c}{RDL} \\ \cline{3-14}
& \multirow[c]{2}{*}{Model} & \multicolumn{2}{c|}{\multirow[c]{2}{*}{LightGBM}} & \multicolumn{2}{c|}{GetML}  & \multicolumn{2}{c||}{Tabular} & \multicolumn{2}{c|}{Linear} & \multicolumn{2}{c|}{ResNet} & \multicolumn{2}{c}{\multirow[c]{2}{*}{\textsc{DBFormer}}} \\
  &  & & & \multicolumn{2}{c|}{XGBoost}  & \multicolumn{2}{c||}{ResNet} & \multicolumn{2}{c|}{SAGE} & \multicolumn{2}{c|}{SAGE}  \\ 
\hline 
 Database & Task & val & test  & val & test & val & test & val & test & val & test & val & test \\ \hline 
\hline

\multicolumn{14}{c}{Binary Classification (AUC ROC $\uparrow$)} \\ \hline

rgastf1 & orig. & .576 & .609 & \bfseries .917 & \bfseries .922 & .535 & .595 & .900 & .897 & .895 & .900 & .904 & .902 \\
expenditures & orig. & .847 & .852 & .813 & .813 & .845 & .847 & \bfseries .920 & \bfseries .917 & .919 & .915 & .893 & .892 \\
\multirow[t]{2}{*}{geneea} & orig. & .994 & .989 & .937 & .953 & .976 & .981 & .986 & .983 & .987 & .975 & \bfseries .996 & \bfseries .994 \\
 & temp. & .990 & .984 & .945 & .934 & .980 & .980 & .986 & .976 & .990 & .974 & \bfseries .998 & \bfseries .990 \\
hepatitis & orig. & .628 & .642 & .964 & .932 & .698 & .633 & \bfseries 1.0 & .997 & \bfseries 1.0 & .996 & \bfseries 1.0 & \bfseries .999 \\
imdb & orig. & .986 & .986 & .549 & .546 & .984 & .985 & \bfseries .993 & \bfseries .993 & \bfseries .993 & \bfseries .993 & \bfseries .993 & \bfseries .993 \\
mondial & orig. & .500 & .500 & NaN & NaN & .500 & .528 & .963 & \bfseries .954 & .971 & .920 & \bfseries .984 & .941 \\
movielens & orig. & .578 & .613 & \bfseries .784 & \bfseries .806 & .536 & .600 & .760 & .789 & .754 & .791 & .755 & .788 \\
musklarge & orig. & .500 & .500 & .619 & \bfseries .760 & .452 & .700 & .876 & .608 & .900 & .632 & \bfseries .905 & .720 \\
musksmall & orig. & .500 & .500 & .800 & .778 & .500 & .583 & .830 & .856 & \bfseries .880 & \bfseries .878 & .760 & .756 \\
mutagenesis & orig. & .901 & .841 & .917 & \bfseries .976 & .845 & .917 & .933 & .845 & .940 & .707 & \bfseries .955 & .812 \\
ncaa & orig. & .562 & .475 & .654 & \bfseries .791 & .568 & .543 & \bfseries .753 & .736 & .720 & .746 & .726 & .712 \\
studentloan & orig. & .500 & .500 & .728 & .753 & .500 & .532 & \bfseries 1.0 & \bfseries 1.0 & \bfseries 1.0 & \bfseries 1.0 & \bfseries 1.0 & \bfseries 1.0 \\
\multirow[t]{2}{*}{amazon} & ichurn & .728 & .730 & NaN & NaN & .652 & .653 & .820 & .824 & \bfseries .822 & \bfseries .825 & .818 & .821 \\
 & uchurn & .570 & .571 & NaN & NaN & .525 & .531 & \bfseries .706 & \bfseries .707 & \bfseries .706 & \bfseries .707 & .703 & .704 \\
\multirow[t]{2}{*}{avito} & clicks & .561 & .542 & NaN & NaN & .536 & .539 & .651 & .642 & \bfseries .653 & \bfseries .671 & .647 & .664 \\
 & visits & .536 & .529 & NaN & NaN & .514 & .508 & .698 & .660 & \bfseries .699 & \bfseries .664 & .694 & .658 \\
\multirow[t]{2}{*}{f1} & dnf & .675 & .681 & .666 & \bfseries .766 & .624 & .716 & .703 & .759 & \bfseries .726 & .753 & .702 & .762 \\
 & top3 & .703 & .745 & .679 & \bfseries .851 & .621 & .748 & \bfseries .727 & .830 & .718 & .831 & .689 & .782 \\
\multirow[t]{2}{*}{stack} & badge & .762 & .740 & .514 & .513 & .688 & .674 & \bfseries .901 & .890 & \bfseries .901 & \bfseries .891 & .897 & .885 \\
 & engmt. & .833 & .823 & .589 & .590 & .788 & .781 & \bfseries .901 & .905 & \bfseries .901 & \bfseries .907 & \bfseries .901 & .904 \\
trial & s.out. & \bfseries .673 & .692 & .626 & \bfseries .699 & .583 & .665 & .624 & .676 & .618 & .692 & .604 & .656 \\
\hline

\multicolumn{14}{c}{Multiclass Classification  (macro F1 score $\uparrow$)} \\ \hline

\multirow[t]{2}{*}{accidents} & orig. & .294 & .292 & .495 & .492 & .347 & .357 & \bfseries 1.0 & \bfseries 1.0 & \bfseries 1.0 & \bfseries 1.0 & \bfseries 1.0 & \bfseries 1.0 \\
 & temp. & .166 & .149 & .319 & .313 & .193 & .180 & .816 & .583 & .819 & .616 & \bfseries .820 & \bfseries .621 \\
craftbeer & orig. & .406 & .232 & .006 & .020 & .394 & \bfseries .308 & .369 & .238 & \bfseries .483 & .275 & .476 & .284 \\
\multirow[t]{2}{*}{dallas} & orig. & .520 & .537 & \bfseries .550 & \bfseries .606 & .460 & .534 & .440 & .467 & .502 & .494 & .394 & .537 \\
 & temp. & .564 & .348 & \bfseries .652 & .347 & .446 & .236 & .417 & \bfseries .411 & .500 & .377 & .464 & .360 \\
diabetes & orig. & .190 & .190 & .402 & .383 & .189 & .190 & .866 & .867 & \bfseries .888 & \bfseries .886 & .880 & .877 \\
\multirow[t]{2}{*}{financial} & orig. & .422 & .416 & \bfseries .935 & \bfseries .886 & .549 & .471 & .483 & .451 & .569 & .621 & .529 & .578 \\
 & temp. & \bfseries .495 & \bfseries .492 & .466 & .466 & .214 & .325 & .292 & .361 & .376 & .362 & .378 & .281 \\
genes & orig. & .060 & .060 & .066 & .100 & .060 & .060 & \bfseries 1.0 & .894 & \bfseries 1.0 & \bfseries .941 & \bfseries 1.0 & .899 \\
hockey & orig. & .626 & .637 & .689 & .725 & .653 & .691 & .692 & \bfseries .727 & \bfseries .697 & .705 & \bfseries .697 & .725 \\
\multirow[t]{2}{*}{legalacts} & orig. & \bfseries .929 & \bfseries .921 & .726 & .723 & .834 & .823 & .835 & .823 & .838 & .827 & .833 & .823 \\
 & temp. & \bfseries .852 & \bfseries .829 & .727 & .695 & .757 & .695 & .787 & .702 & .778 & .700 & .775 & .705 \\
premiere\- & orig. & .270 & .250 & \bfseries .630 & \bfseries .659 & .417 & .400 & .594 & .438 & .524 & .385 & .452 & .362 \\
league & temp. & .250 & .222 & \bfseries .626 & \bfseries .667 & .467 & .407 & .531 & .521 & .572 & .425 & .467 & .470 \\
tpcd & orig. & .183 & .176 & .193 & .196 & .156 & .168 & \bfseries .738 & \bfseries .743 & .573 & .579 & .691 & .693 \\
webkp & orig. & .139 & .130 & \bfseries .350 & \bfseries .366 & .139 & .130 & .263 & .238 & .294 & .209 & .284 & .273 \\ \hline \hline
\multirow[c]{2}{*}{Avg. Rank} &  & \multicolumn{2}{c|}{4.49} & \multicolumn{2}{c|}{3.61} & \multicolumn{2}{c|}{4.71} & \multicolumn{2}{c|}{2.75} & \multicolumn{2}{c|}{\bfseries 2.66} & \multicolumn{2}{c}{2.79} \\ \cline{3-14}
& & \multicolumn{2}{c|}{2.91} &\multicolumn{2}{c|}{2.41} & \multicolumn{2}{c|}{3.16} & \multicolumn{6}{c}{\bfseries 1.53}  \\
\hline
\end{tabular}

%% file: tables/class-tabular-corrected.tex
\begin{tabularx}{\textwidth}{lC{3}||C{2}C{2}|C{2}C{2}|C{2}C{2}|C{2}C{2}|C{2}C{2}|C{2}C{2}}
& \multirow[c]{2}{*}{Model} &  \multicolumn{4}{C{8}|}{LightGBM} &  \multicolumn{4}{C{8}|}{ResNet Tabular}  & \multicolumn{4}{C{8}}{RDL} \\ \cline{3-14}
&  & \multicolumn{2}{C{4.5}|}{Base} & \multicolumn{2}{C{4}|}{Joined}  & \multicolumn{2}{C{4}|}{Base} & \multicolumn{2}{C{4}|}{Joined} & \multicolumn{2}{C{4}|}{2 layer} & \multicolumn{2}{C{4}}{Best} \\
\hline 
 Database & Task & val & test  & val & test & val & test & val & test & val & test & val & test \\ \hline 

 \multicolumn{14}{c}{Binary Classification} \\ \hline
 
ergastf1 & orig. & .576 & .609 & \bfseries \underline{.913} & \bfseries \underline{.906} & .535 & .595 & \bfseries .873 & \bfseries .893 & .900 & .897 & .904 & .902 \\
hepatitis & orig. & .628 & .642 & \bfseries .920 & \bfseries .904 & .698 & .633 & .696 & .634 & .998 & .987 & 1.0 & .997 \\
mondial & orig. & .500 & .500 & \bfseries .942 & \bfseries .855 & .500 & .528 & \bfseries .920 & \bfseries .917 & .963 & .935 & .984 & .941 \\
movielens & orig. & .578 & .613 & .571 & .613 & .536 & .600 & .536 & .600 & .551 & .618 & .760 & .789 \\
ncaa & orig. & .562 & .475 & \bfseries .742 & \bfseries .729 & .568 & .543 & \bfseries .664 & \bfseries \underline{.776} & .672 & .764 & .753 & .736 \\
studentloan & orig. & .500 & .500 & \bfseries \underline{1.0} & \bfseries \underline{1.0} & .500 & .532 & \bfseries \underline{1.0} & \bfseries \underline{1.0} & 1.0 & 1.0 & 1.0 & 1.0 \\
amazon & uchurn & .570 & .571 & .561 & .563 & .525 & .531 & .525 & .531 & .699 & .699 & .706 & .707 \\
\multirow[t]{2}{*}{avito} & clicks & .561 & .542 & .562 & .550 & .536 & .539 & .535 & .538 & .623 & .647 & .653 & .671 \\
 & visits & .536 & .529 & .535 & .532 & .514 & .508 & .514 & .509 & .680 & .655 & .699 & .664 \\
stack & badge & .762 & .740 & .778 & .754 & .688 & .674 & .687 & .675 & .892 & .880 & .901 & .890 \\

\hline
\multicolumn{14}{c}{Multiclass Classification} \\ \hline

\multirow[t]{2}{*}{accidents} & orig. & .294 & .292 & \bfseries .369 & \bfseries .367 & .347 & .357 & .359 & .362 & 1.0 & 1.0 & 1.0 & 1.0 \\
 & temp. & .166 & .149 & .166 & .152 & .193 & .180 & .188 & .177 & .801 & .671 & .820 & .621 \\
craftbeer & orig. & .406 & .232 & .448 & .271 & .394 & .308 & .394 & .308 & .442 & .335 & .483 & .275 \\
diabetes & orig. & .190 & .190 & .190 & .190 & .189 & .190 & .189 & .190 & .871 & .869 & .888 & .886 \\
genes & orig. & .060 & .060 & .060 & .060 & .060 & .060 & .060 & .060 & 1.0 & .898 & 1.0 & .894 \\
premiere\-& orig. & .270 & .250 & \bfseries .370 & \bfseries .306 & .417 & .400 & .438 & .349 & .570 & .409 & .594 & .438 \\
league & temp. & .250 & .222 & \bfseries .427 & \bfseries .339 & .467 & .407 & .442 & .402 & .523 & .419 & .572 & .425 \\
tpcd & orig. & .183 & .176 & \bfseries \underline{.822} & \bfseries \underline{.821} & .156 & .168 & \bfseries .671 & \bfseries .667 & .600 & .603 & .738 & .743 \\
webkp & orig. & .139 & .130 & .139 & .130 & .139 & .130 & .139 & .130 & .233 & .211 & .294 & .209 \\
\hline
\end{tabularx}

%% file: tables/database-characteristics-corrected.tex
\begin{tabularx}{\textwidth}{l||C{2.9}|C{2.9}C{2.9}C{2.9}|C{2.9}C{2.9}C{2.9}|C{2.9}C{2.9}C{2.9}}
 \multirow[c]{2}{*}{Feature} & Base & \multicolumn{3}{C{8.7}|}{RDL} & \multicolumn{3}{C{8.7}|}{Propositional.} & \multicolumn{3}{C{8.7}}{Tabular Learning} \\ \cline{2-11}
  & Med. & Q1 & Median & Q3 & Q1 & Median & Q3 & Q1 & Median & Q3 \\

\hline
\#Tab. & 6 & 3 & 7 & 8 & 3 & 7 & 9 & 3.5 & 4 & 5 \\
\#FK & 6 & 2 & 6 & 12 & 3 & 6 & 13 & 3.25 & 4 & 5 \\
\#Factual & 35 & 18 & 26 & 48 & 26 & 47 & 110 & 23.25 & 28 & 32.75 \\
\#Cat. & 11 & 3 & 11 & 18 & 14 & 17 & 88 & 10.25 & 13 & 14 \\
\#Num. & 6 & 1 & 6 & 11 & 3 & 5 & 12 & 3 & 3 & 6.75 \\
\#Text & 6 & 2 & 6 & 14 & 0 & 2 & 9 & 2.5 & 3 & 3 \\
\#Time & 2 & 0 & 1 & 4 & 0 & 1 & 7 & 3.75 & 5 & 5 \\
\#Rows & 472k & 21k & 1.5M & 5.6M & 11k & 98k & 546k & 810k & 1.4M & 1.8M \\
\#Links & 752k & 43k & 2.2M & 8.2M & 32k & 228k & 1.1M & 814k & 1.7M & 2.3M \\
\hline

Diameter & 3 & 2 & 2 & 4 & 2 & 3 & 3 & 2.5 & 3 & 3 \\
1-to-1 & 0 & 0 & 0 & 6 & 0 & 0 & 0 & 0 & 0 & 1.5 \\
1-to-M & 5 & 2 & 3 & 11 & 3 & 5 & 13 & 3.25 & 4 & 4.25 \\
\hline

\#Train & 12k & 0.8k & 59k & 654k & 0.2k & 0.5k & 5k & 0.5k & 226k & 455k \\
\#T.Factual & 5 & 1 & 5 & 13 & 2 & 5 & 6 & 4.25 & 12.5 & 20 \\
T.Cat. & 2 & 0 & 0 & 4 & 1 & 2 & 4 & 0.75 & 5.5 & 10 \\
T.Num. & 0 & 0 & 0 & 1 & 0 & 0 & 2 & 1.5 & 2.5 & 3 \\
T.Text & 1 & 0 & 1 & 4 & 0 & 0 & 3 & 1.5 & 2 & 2 \\
T.Time & 1 & 0 & 0 & 1 & 0 & 1 & 1 & 0.75 & 3 & 5 \\
\hline

Eccentri. & 10.64 & 7.92 & 12.24 & 17.84 & 9.36 & 13.90 & 19.20 & 4.59 & 7.74 & 9.92 \\
Density & 0.025 & 0.009 & 0.022 & 0.107 & 0.001 & 0.002 & 0.028 & 0.067 & 0.109 & 0.243 \\
\hline
\end{tabularx}

%% file: tables/tab-like.tex
\begin{tabularx}{\textwidth}{L{7}||C{4}|C{4}|C{3.5}|C{5}|C{5}|C{5.5}}
Database & Domain & \#Tables & \#FK & \#Factual & one-to-one & many-to-one \\ 
\hline
atherosclerosis & Med. & 4 & 3 & 191 & 2 & 1 \\
bupa & Med. & 9 & 8 & 16 & 7 & 1 \\
cde & Gov. & 3 & 2 & 87 & 2 & 0 \\
pima & Med. & 9 & 8 & 9 & 8 & 0 \\
satellite & Indstr. & 34 & 34 & 67 & 34 & 0 \\
voc & Retail & 8 & 7 & 89 & 6 & 1 \\
\end{tabularx}

%% file: tables/graph-like.tex
\begin{tabularx}{\textwidth}{L{7}||C{4}|C{4}|C{3.5}|C{5}|C{5}|C{5.5}}
Database & Domain & \#Tables & \#FK & \#Factual & one-to-one & many-to-one \\ 
\hline
carcinogenesis & Med. & 6 & 13 & 4 & 2 & 11 \\
cora & Edu. & 3 & 3 & 2 & 0 & 3 \\
mesh & Industry & 29 & 33 & 37 & 24 & 9 \\
pima & Med. & 9 & 8 & 9 & 8 & 0 \\
toxicology & Med. & 4 & 5 & 3 & 0 & 5 \\
\end{tabularx}

%% file: tables/db-info.tex
\begin{tabularx}{0.95\textwidth}{l||C{4}|C{3}|C{3}|C{3}|C{3}|C{3}|C{3}|C{3}}
Database & Dom. & \#Tab. & \#FK & \#Fact.  & Diam. & Cycle & 1:1 & 1:N \\ 
\hline \hline
\multicolumn{9}{c}{CTU Relational Databases} \\ \hline
accidents & Gov. & 3 & 3 & 38 & 1 & \cmark & 0 & 3 \\
basketballmen & Sport & 9 & 9 & 187 & 5 & \cmark & 2 & 8 \\
biodegradability & Med. & 5 & 5 & 6 & 3 & \cmark & 0 & 5 \\
countries & Geo. & 3 & 2 & 63 & 2 & \xmark & 0 & 2 \\
craftbeer & Entmt. & 2 & 1 & 8 & 1 & \xmark & 0 & 1 \\
dallas & Gov. & 3 & 2 & 24 & 2 & \xmark & 0 & 2 \\
diabetes & Edu. & 3 & 2 & 4 & 2 & \xmark & 0 & 2 \\
ergastf1 & Sport & 13 & 19 & 82 & 3 & \cmark & 0 & 19 \\
expenditures & Retail & 3 & 2 & 19 & 2 & \xmark & 0 & 2 \\
financial & Fin. & 8 & 8 & 39 & 3 & \cmark & 6 & 5 \\
fnhk & Med. & 3 & 2 & 21 & 2 & \xmark & 0 & 2 \\
geneea & Gov. & 19 & 20 & 99 & 7 & \cmark & 8 & 16 \\
genes & Med. & 3 & 3 & 11 & 2 & \cmark & 0 & 3 \\
grants & Edu. & 12 & 11 & 30 & 4 & \xmark & 2 & 10 \\
hepatitis & Med. & 7 & 6 & 16 & 4 & \xmark & 6 & 3 \\
hockey & Sport & 19 & 27 & 273 & 4 & \cmark & 8 & 23 \\
imdb & Entmt. & 7 & 6 & 12 & 5 & \xmark & 0 & 6 \\
lahman & Sport & 25 & 31 & 319 & 6 & \cmark & 6 & 28 \\
legalacts & Gov. & 4 & 4 & 24 & 3 & \cmark & 0 & 4 \\
mondial & Geo. & 33 & 62 & 125 & 5 & \cmark & 12 & 55 \\
movielens & Entmt. & 7 & 6 & 14 & 4 & \xmark & 0 & 6 \\
musklarge & Med. & 2 & 1 & 167 & 1 & \xmark & 0 & 1 \\
musksmall & Med. & 2 & 1 & 167 & 1 & \xmark & 0 & 1 \\
mutagenesis & Med. & 3 & 3 & 9 & 2 & \cmark & 0 & 3 \\
ncaa & Sport & 8 & 15 & 99 & 3 & \cmark & 0 & 15 \\
premiereleague & Sport & 4 & 5 & 209 & 2 & \cmark & 0 & 5 \\
restbase & Retail & 3 & 3 & 10 & 1 & \cmark & 2 & 2 \\
seznam & Retail & 4 & 3 & 10 & 2 & \xmark & 0 & 3 \\
sfscores & Gov. & 3 & 2 & 22 & 2 & \xmark & 0 & 2 \\
stats & Edu. & 8 & 13 & 50 & 3 & \cmark & 2 & 12 \\
studentloan & Fin. & 10 & 9 & 15 & 3 & \xmark & 12 & 3 \\
tpcd & Retail & 8 & 8 & 48 & 4 & \cmark & 1 & 7 \\
triazine & Med. & 2 & 1 & 13 & 1 & \xmark & 0 & 1 \\
walmart & Retail & 4 & 3 & 27 & 3 & \xmark & 0 & 3 \\
webkp & Edu. & 3 & 3 & 5 & 2 & \cmark & 0 & 3 \\
\hline \multicolumn{9}{c}{\textsc{RelBench} Databases} \\ \hline
amazon & Retail & 3 & 2 & 10 & 2 & \xmark & 0 & 2 \\
avito & Retail & 8 & 11 & 23 & 2 & \cmark & 0 & 11 \\
f1 & Sport & 9 & 13 & 45 & 3 & \cmark & 0 & 13 \\
stack & Edu. & 7 & 12 & 33 & 3 & \cmark & 2 & 11 \\
trial & Med. & 15 & 15 & 110 & 4 & \cmark & 4 & 13 \\
\hline \hline
\end{tabularx}

%% file: bibliography.bib
@book{vsir2021deep,
  title={Deep Learning with Relational Logic Representations},
  author={{\v{S}}{\'\i}r, Gustav},
  year={2021},
  publisher={Czech Technical University}
}

@inproceedings{vogel2024wikidbs,
	title = {{WikiDBs}: {A} {Large}-{Scale} {Corpus} {Of} {Relational} {Databases} {From} {Wikidata}},
	volume = {37},
	booktitle = {Advances in {Neural} {Information} {Processing} {Systems}},
	publisher = {Curran Associates, Inc.},
	author = {Vogel, Liane and Bodensohn, Jan-Micha and Binnig, Carsten},
	editor = {Globerson, A. and Mackey, L. and Belgrave, D. and Fan, A. and Paquet, U. and Tomczak, J. and Zhang, C.},
	year = {2024},
	pages = {41186--41201},
}

@inproceedings{zahradnik2023deep,
    title={A Deep Learning Blueprint for Relational Databases},
    author={Luk{\'a}{\v{s}} Zahradn{\'\i}k and Jan Neumann and Gustav {\v{S}}{\'\i}r},
    booktitle={NeurIPS 2023 Second Table Representation Learning Workshop},
    year={2023}
}

@article{zhang2023gfs,
  title={Gfs: Graph-based feature synthesis for prediction over relational databases},
  author={Zhang, Han and Gan, Quan and Wipf, David and Zhang, Weinan},
  journal={arXiv preprint arXiv:2312.02037},
  year={2023}
}

@inproceedings{
robinson2024relbench,
title={RelBench: A Benchmark for Deep Learning on Relational Databases},
author={Joshua Robinson and Rishabh Ranjan and Weihua Hu and Kexin Huang and Jiaqi Han and Alejandro Dobles and Matthias Fey and Jan Eric Lenssen and Yiwen Yuan and Zecheng Zhang and Xinwei He and Jure Leskovec},
booktitle={The Thirty-eight Conference on Neural Information Processing Systems Datasets and Benchmarks Track},
year={2024}
}

@inproceedings{
    fey2024position,
    title={Position: Relational Deep Learning - Graph Representation Learning on Relational Databases},
    author={Matthias Fey and Weihua Hu and Kexin Huang and Jan Eric Lenssen and Rishabh Ranjan and Joshua Robinson and Rex Ying and Jiaxuan You and Jure Leskovec},
    booktitle={Forty-first International Conference on Machine Learning},
    year={2024}
}

@article{borisov2022deep,
  title={Deep neural networks and tabular data: A survey},
  author={Borisov, Vadim and Leemann, Tobias and Se{\ss}ler, Kathrin and Haug, Johannes and Pawelczyk, Martin and Kasneci, Gjergji},
  journal={IEEE transactions on neural networks and learning systems},
  year={2022},
  publisher={IEEE}
}

@inbook{Lavrač2021,
	title        = {Machine Learning Background},
	author       = {Lavra{\v{c}}, Nada and Podpe{\v{c}}an, Vid and Robnik-{\v{S}}ikonja, Marko},
	year         = 2021,
	booktitle    = {Representation Learning: Propositionalization and Embeddings},
	publisher    = {Springer International Publishing},
	address      = {Cham},
	pages        = {17--53},
	isbn         = {978-3-030-68817-2},
}

@misc{Cvitkovic2020,
	title        = {Supervised Learning on Relational Databases with Graph Neural Networks},
	author       = {Milan Cvitkovic},
	year         = 2020,
	eprint       = {2002.02046},
	archiveprefix = {arXiv},
	primaryclass = {cs.LG}
}

@article{Codd1970,
	title        = {A relational model of data for large shared data banks},
	author       = {Codd, E. F.},
	year         = 1970,
	journal      = {Commun. ACM},
	publisher    = {Association for Computing Machinery},
	address      = {New York, NY, USA},
	volume       = 13,
	number       = 6,
	pages        = {377–387},
	issn         = {0001-0782},
	issue_date   = {June 1970},
	numpages     = 11
}

@article{Kingma2014,
	title        = {Adam: A method for stochastic optimization},
	author       = {Kingma, Diederik P and Ba, Jimmy},
	year         = 2014,
	journal      = {arXiv preprint arXiv:1412.6980}
}

@misc{chen2023trompt,
	title        = {Trompt: Towards a Better Deep Neural Network for Tabular Data},
	author       = {Kuan-Yu Chen and Ping-Han Chiang and Hsin-Rung Chou and Ting-Wei Chen and Tien-Hao Chang},
	year         = 2023,
	eprint       = {2305.18446},
	archiveprefix = {arXiv},
	primaryclass = {cs.LG}
}

@misc{somepalli2021saint,
	title        = {SAINT: Improved Neural Networks for Tabular Data via Row Attention and Contrastive Pre-Training},
	author       = {Gowthami Somepalli and Micah Goldblum and Avi Schwarzschild and C. Bayan Bruss and Tom Goldstein},
	year         = 2021,
	eprint       = {2106.01342},
	archiveprefix = {arXiv},
	primaryclass = {cs.LG}
}

@inproceedings{arik2021tabnet,
  title={Tabnet: Attentive interpretable tabular learning},
  author={Arik, Sercan {\"O} and Pfister, Tomas},
  booktitle={Proceedings of the AAAI conference on artificial intelligence},
  pages={6679--6687},
  year={2021}
}

@software{getml,
	title        = {getML},
	author       = {{Code17 GmbH}},
	url          = {https://getml.com},
	date         = 2024
}

@inproceedings{cropper2020turning30,
  title     = {Turning 30: New Ideas in Inductive Logic Programming},
  author    = {Cropper, Andrew and Dumančić, Sebastijan and Muggleton, Stephen H.},
  booktitle = {Proceedings of the Twenty-Ninth International Joint Conference on
               Artificial Intelligence, {IJCAI-20}},
  editor    = {Christian Bessiere},
  pages     = {4833--4839},
  year      = {2020},
  
}

@inproceedings{velivckovic2018graph,
  title={Graph Attention Networks},
  author={Veli{\v{c}}kovi{\'c}, Petar and Cucurull, Guillem and Casanova, Arantxa and Romero, Adriana and Li{\`o}, Pietro and Bengio, Yoshua},
  booktitle={International Conference on Learning Representations},
  year={2018}
}

@inproceedings{chen2016xgboost,
  title={Xgboost: A scalable tree boosting system},
  author={Chen, Tianqi and Guestrin, Carlos},
  booktitle={Proceedings of the 22nd acm sigkdd international conference on knowledge discovery and data mining},
  pages={785--794},
  year={2016}
}

@book{codd1990relational,
  title={The relational model for database management: version 2},
  author={Codd, Edgar F},
  year={1990},
  publisher={Addison-Wesley Longman Publishing Co., Inc.}
}

@article{kramer2001propositionalization,
  title={Propositionalization approaches to relational data mining},
  author={Kramer, Stefan and Lavra{\v{c}}, Nada and Flach, Peter},
  journal={Relational data mining},
  pages={262--291},
  year={2001},
  publisher={Springer}
}

@article{motl2015ctu,
  title={The CTU Prague Relational Learning Repository},
  author={Motl, Jan and Schulte, Oliver},
  journal={arXiv preprint arXiv:1511.03086},
  year={2015}
}

@article{hu2020tabtransformer,
  title={TabTransformer: Tabular Data Modeling Using Contextual Embeddings},
  author={Hu, Xuanwei and Tang, Weijing and Hsieh, Cheng-Kang and Shi, Shuaiwen},
  journal={arXiv preprint arXiv:2012.06678},
  year={2020}
}

@article{vaswani2017attention,
  title={Attention is all you need},
  author={Vaswani, Ashish and Shazeer, Noam and Parmar, Niki and Uszkoreit, Jakob and Jones, Llion and Gomez, Aidan N and Kaiser, {\L}ukasz and Polosukhin, Illia},
  journal={Advances in neural information processing systems},
  
  year={2017}
}

@book{gallier2015logic,
	title        = {Logic for computer science: foundations of automatic theorem proving},
	author       = {Gallier, Jean H},
	year         = 2015,
	publisher    = {Courier Dover Publications}
}

@article{wu2020comprehensive,
	title        = {A comprehensive survey on graph neural networks},
	author       = {Wu, Zonghan and Pan, Shirui and Chen, Fengwen and Long, Guodong and Zhang, Chengqi and Philip, S Yu},
	year         = 2020,
	journal      = {IEEE Transactions on Neural Networks and Learning Systems},
	publisher    = {IEEE}
}

@inproceedings{hamilton2017inductive,
	title        = {Inductive representation learning on large graphs},
	author       = {Hamilton, Will and Ying, Zhitao and Leskovec, Jure},
	year         = 2017,
	booktitle    = {Advances in neural information processing systems},
	pages        = {1024--1034}
}

@article{muggleton1994inductive,
	title        = {Inductive logic programming: Theory and methods},
	author       = {Muggleton, Stephen and De Raedt, Luc},
	year         = 1994,
	journal      = {The Journal of Logic Programming},
	publisher    = {Elsevier},
	volume       = 19
}

@book{Raedt,
	title        = {Logical and Relational Learning},
	author       = {{De Raedt}, Luc},
	year         = 2008,
	publisher    = {Springer}
}

@book{propos,
	title        = {Comparative evaluation of approaches to propositionalization},
	author       = {Krogel, Mark-A and Rawles, Simon and {\v{Z}}elezn{\'y}, Filip and Flach, Peter A and Lavra{\v{c}}, Nada and Wrobel, Stefan},
	year         = 2003,
	publisher    = {Springer}
}

@inproceedings{chamberlin_sequel_1974,
    address = {New York, NY, USA},
    series = {{SIGFIDET} '74},
    title = {{SEQUEL}: {A} structured {English} query language},
    isbn = {978-1-4503-7415-6},
    shorttitle = {{SEQUEL}},
    urldate = {2024-11-12},
    booktitle = {Proceedings of the 1974 {ACM} {SIGFIDET} (now {SIGMOD}) workshop on {Data} description, access and control},
    publisher = {Association for Computing Machinery},
    author = {Chamberlin, Donald D. and Boyce, Raymond F.},
    year = {1974},
    pages = {249--264},
}

@article{peleska_transformers_2024,
  title={Transformers Meet Relational Databases},
  author={Pele{\v{s}}ka, Jakub and {\v{S}}{\'\i}r, Gustav},
  journal={arXiv preprint arXiv:2412.05218},
  year={2024}
}

@incollection{sqlalchemy,
  place={Mountain View},
  title={SQLAlchemy},
  booktitle={The Architecture of Open Source Applications Volume II: Structure, Scale, and a Few More Fearless Hacks},
  publisher={aosabook.org},
  author={Bayer, Michael},
  year={2012}
}

@article{rossi2020temporal,
  title={Temporal graph networks for deep learning on dynamic graphs},
  author={Rossi, Emanuele and Chamberlain, Ben and Frasca, Fabrizio and Eynard, Davide and Monti, Federico and Bronstein, Michael},
  journal={arXiv preprint arXiv:2006.10637},
  year={2020}
}

@inproceedings{gorishniy_revisiting_2021,
    address = {Red Hook, NY, USA},
    series = {{NIPS} '21},
    title = {Revisiting deep learning models for tabular data},
    isbn = {978-1-71384-539-3},
    urldate = {2025-03-12},
    booktitle = {Proceedings of the 35th {International} {Conference} on {Neural} {Information} {Processing} {Systems}},
    publisher = {Curran Associates Inc.},
    author = {Gorishniy, Yury and Rubachev, Ivan and Khrulkov, Valentin and Babenko, Artem},
    year = {2021},
    pages = {18932--18943},
}

@inproceedings{brody2022how,
title={How Attentive are Graph Attention Networks? },
author={Shaked Brody and Uri Alon and Eran Yahav},
booktitle={International Conference on Learning Representations},
year={2022}
}

@article{liang2024foundations,
  title={Foundations \& trends in multimodal machine learning: Principles, challenges, and open questions},
  author={Liang, Paul Pu and Zadeh, Amir and Morency, Louis-Philippe},
  journal={ACM Computing Surveys},
  
  number={10},
  pages={1--42},
  year={2024},
  publisher={ACM New York, NY}
}

@inproceedings{hu2021unit,
  title={Unit: Multimodal multitask learning with a unified transformer},
  author={Hu, Ronghang and Singh, Amanpreet},
  booktitle={Proceedings of the IEEE/CVF international conference on computer vision},
  pages={1439--1449},
  year={2021}
}

@inproceedings{ke_lightgbm_2017,
    address = {Red Hook, NY, USA},
    series = {{NIPS}'17},
    title = {{LightGBM}: a highly efficient gradient boosting decision tree},
    isbn = {978-1-5108-6096-4},
    shorttitle = {{LightGBM}},
    urldate = {2025-03-14},
    booktitle = {Proceedings of the 31st {International} {Conference} on {Neural} {Information} {Processing} {Systems}},
    publisher = {Curran Associates Inc.},
    author = {Ke, Guolin and Meng, Qi and Finley, Thomas and Wang, Taifeng and Chen, Wei and Ma, Weidong and Ye, Qiwei and Liu, Tie-Yan},
    year = {2017},
    pages = {3149--3157},
}

@article{chen_relgnn_2025,
    title = {{RelGNN}: {Composite} {Message} {Passing} for {Relational} {Deep} {Learning}},
    shorttitle = {{RelGNN}},
    urldate = {2025-02-17},
    author = {Chen, Tianlang and Kanatsoulis, Charilaos and Leskovec, Jure},
    
    year = {2025},
    journal={arXiv preprint arXiv:2502.06784}
}

@article{natekin2013gradient,
  title={Gradient boosting machines, a tutorial},
  author={Natekin, Alexey and Knoll, Alois},
  journal={Frontiers in neurorobotics},
  volume={7},
  pages={21},
  year={2013},
  publisher={Frontiers Media SA}
}

@article{wang_4dbinfer_2024,
  title={4DBInfer: A 4d benchmarking toolbox for graph-centric predictive modeling on RDBs},
  author={Wang, Minjie and Gan, Quan and Wipf, David and Zhang, Zheng and Faloutsos, Christos and Zhang, Weinan and Zhang, Muhan and Cai, Zhenkun and Li, Jiahang and Mao, Zunyao and others},
  journal={Advances in Neural Information Processing Systems},
  volume={37},
  pages={27236--27273},
  year={2024}
}

@book{hamilton_graph_2020,
    title = {Graph {Representation} {Learning}},
    isbn = {978-1-68173-964-9},
    language = {en},
    publisher = {Morgan \& Claypool Publishers},
    author = {Hamilton, William L.},
    
    year = {2020},
}

@inproceedings{maali_enabling_2010,
    address = {Berlin, Heidelberg},
    title = {Enabling {Interoperability} of {Government} {Data} {Catalogues}},
    isbn = {978-3-642-14799-9},
    language = {en},
    booktitle = {Electronic {Government}},
    publisher = {Springer},
    author = {Maali, Fadi and Cyganiak, Richard and Peristeras, Vassilios},
    editor = {Wimmer, Maria A. and Chappelet, Jean-Loup and Janssen, Marijn and Scholl, Hans J.},
    year = {2010},
    pages = {339--350},
}

@article{white_pubmed_2020,
    title = {{PubMed} 2.0},
    volume = {39},
    issn = {0276-3869},
    number = {4},
    urldate = {2025-03-15},
    journal = {Medical Reference Services Quarterly},
    author = {White, Jacob},
    
    year = {2020},
    pmid = {33085945},
    pages = {382--387},
}
